\def\year{2015}
\documentclass[letterpaper]{article}
\usepackage{aaai}
\usepackage{times}
\usepackage{helvet}
\usepackage{courier}
\usepackage{url}

\usepackage{graphicx}
\usepackage{amsmath}

\begin{document}
%
\title{Poker-CNN: A Pattern Learning Strategy for Making Draws and Bets \\
in Poker Games}




\author{
Nikolai Yakovenko \\
Columbia University, New York \\
\texttt{nvy2101@columbia.edu} \\
\And
Liangliang Cao \\
Columbia University and Yahoo Labs, New York \\
\texttt{liangliang.cao@gmail.com} \\
\AND
Colin Raffel \\
Columbia University, New York \\
\texttt{craffel@gmail.com} \\
\And
James Fan \\
Columbia University, 
New York \\
\texttt{jfan.us@gmail.com} \\
}

%


\maketitle

\begin{abstract}

Poker is a family of card games that includes many variations. We hypothesize that most poker games can be solved as a pattern matching problem, and propose creating a strong poker playing system based on a unified poker representation. Our poker player learns through iterative self-play, and improves its understanding of the game by training on the results of its previous actions without sophisticated domain knowledge. We evaluate our system on three poker games: single player video poker, two-player Limit Texas Hold'em, and finally two-player 2-7 triple draw  poker. We show that our model can quickly learn  patterns in these very different poker games while it improves from zero knowledge to a competitive player against human experts.

The contributions of this paper include: (1) a novel representation for poker games, extendable to different poker variations, (2) a CNN based learning model that can effectively learn the patterns in three different games, and (3) a self-trained system that significantly beats the heuristic-based program on which it is trained, and our system is competitive against human expert players.

\end{abstract}

\section{Introduction}

Poker, a family of different card games with different rules,
is a challenging problem for artificial intelligence because the game state space is huge and the numerous variations require different domain knowledge. 

There have been a number of efforts to design an AI system for poker games over the past few years, focusing on the Texas Hold'em style of poker~\cite{rubin2011computer} \cite{poker-science15}. They have achieved competitive performance against the world’s best heads-up No Limit Texas Hold'em players~\cite{rutkin2015ai}, and found an effective equilibrium solution for heads-up Limit Texas Hold'em\cite{cfr-bowling2015heads}.
These systems treat poker as 
imperfect-information games, and develop search strategies specific to Texas Hold'em to find the game equilibrium, with various abstractions to group similar game states together, thus reducing the complexity of the game. However, equilibrium-finding requires domain knowledge and considerable effort to condense a particular variation of poker's massive number of possible states into a more manageable set of similar states. As a result, existing poker playing systems are difficult to build, and their success is limited to a specific game, e.g., heads-up limit Texas Hold'em \cite{poker-science15}. In contrast, human poker players show more general expertise, as they are usually good at many poker variations.



We classify previous works  \cite{rubin2011computer} \cite{poker-science15} as {\em expert knowledge based approaches}. This paper proposes a {\em data driven approach} which uses the same machinery to learn patterns for different poker games from training data produced by inferior simple heuristic players, as well as from self-play. The training data are game histories, and the output is a neural network that estimates the gain/loss of each legal game action in the current game context. We believe that this data-driven approach can be applied to a wide variety of poker games with little game-specific knowledge. 



However, there are several challenges in learning from these synthesized training data:



\begin{itemize}
	\item Representation. A learning model requires a good representation which includes rich information present in different poker games, such as the player's private cards, any known public information, and the actions taken on previous rounds of the game. 
	\item Sophistication. The training data obtained from simulation is far from perfect. Our training data first came from the simple heuristic players, and they are not  competitive enough for serious poker players. 
    
\end{itemize}

To solve these challenges, we employ a number of techniques. We propose a novel representation for poker games, which can be used across different poker games with different rules to capture the card information as well as betting sequences. Based on this representation, we develop a deep neural network which can learn the knowledge for each poker game, and embed such knowledge into its parameters. 
By acknowledging that the training data is not perfect, we propose to improve our  network by learning on its own. At every step, our network can be used to simulate more data. As a result, our trained model is always more competitive than the model generating the training data.




We have evaluated our  system on three poker variations. 
Our Poker-CNN obtains an average return of $\$0.977\pm 0.005$ in video poker, which is similar to human  players' performance. For Texas Hold'em, Poker-CNN beats an open source poker agent by a large margin, and it statistically ties with a professional human player over a 500 hand sample. Poker-CNN for the 2-7 triple draw poker game plays significantly better than the heuristic model that it trained on, it performed competitively against a human expert, and was even ahead of that human expert over the first 100 hands~\footnote{Source code will be available soon}.

\section{Related Work}

There have been many works in the domain of poker play systems. For example, 
Bowling~\cite{cfr-bowling2015heads} et al. have focused a great deal of research on heads-up limit Texas Hold'em, and recently claimed that this limited game is now essentially weakly solved
. 
Their works employ a method called Counterfactual Regret minimization (CFR) to find an equilibrium solution for heads-up Limit Texas Holdem, and which explores
all possible states in the poker game. 
Most existing work on CFR is applied to Texas Hold'em. 

While CFR-based approaches have achieved breakthroughs in Texas Hold'em, there are limitations to adapting them to other poker games. First, it is not easy or straightforward to generalize these works to other poker variations. Second, because quite a few poker games have a  search space larger than heads-up limit Texas Hold'em, it is very difficult to traverse the game states and find the best responses \cite{cfr-zinkevich2007regret} \cite{cfr-johanson2011accelerating} \cite{cfr-johanson2013evaluating} for those games. 
For example, 2-7 triple draw has 7 rounds of action instead of 4, and 5 hidden cards instead of 2 as in limit Texas Hold'em. The number of states is orders of magnitudes larger. It may not always be practical to create an abstraction to use the CFR algorithm, and to find a Nash equilibrium solution with a reasonable amount of computional resources (at least without ignoring some game-state information).



The Carnegie Melon team's winning No Limit Hold'em entry recently challenged some of the best human players in the world, in a long two-player No Limit Hold'em match. Human experts won the match, but not by a definitive margin of error, despite playing 80,000 hands over two weeks \cite{sandholm2015abstraction}.




Most previous works on poker are limited to Texas Hold'em. 
This paper outlines  a data-driven approach for general poker games.  We propose a new representation for poker games, as well as an accompanying way of generating huge amounts of training samples by self-playing and iterative training. Our model is motivated by the recent success of 
Convolutional Neural Networks (CNNs) for large scale learning \cite{fukushima1980neocognitron} \cite{LeCun-PIEEE-1998},  \cite{DBLP:conf/nips/KrizhevskySH12} \cite{deepface-cvpr14}. Similar to the recent progress for the board game go \cite{clark2014teaching,maddison2014move}, and in reinforcement learning on video games \cite{mnih2015human}, we find that CNN works well for poker, and that it outperforms traditional neural network models such as fully-connected networks. Note that our approach is different from the works in \cite{dahl2001reinforcement} \cite{tesauro1994td} 
since our network does not assume expert knowledge of the poker game.

\section{Representation}

In order to create a system that can play many variations of poker, we need to create a representation framework that can encode the state-space in any poker game. In this paper, we show how a 3D tensor based representation can represent three poker games: video poker, heads-up limit Texas Hold'em and heads-up limit 2-7 triple draw.

\subsection{Three poker games}



\begin{enumerate}
	\item Video poker: video poker is a popular single player single round game played in casinos all over the world. A player deposits \$1, and is dealt five random cards. He can keep any or all of the five cards, and the rest are replaced by new cards from the deck. The player's earning is based on his final hand and a pay out table (e.g. a pair of kings earns \$1). Video poker is a simple game with a single player move that has 32 possible choices. 
    
    \item Texas Hold'em: Texas Hold'em is a multi-player game with four betting rounds. Two cards (hole cards) are dealt face down to each player and then five community cards are placed face-up by the dealer in three rounds - first three cards ("the flop") then an additional single card ("the turn" or "fourth street") finally another additional card ("the river" or "fifth street"). The best five card poker hand from either the community or their hole cards wins. Players have the option to check, bet, raise or fold after each deal. 
    
    \item 2-7 triple draw poker: Triple draw is also a multi-round multi-player game. Each player is dealt five cards face down, and they can make three draws from the deck, with betting rounds in between. In each drawing round, players choose to replace any or all cards with new cards from the deck. All cards are face-down, but players can observe how many cards his opponent has exchanged. Triple draw combines both the betting of Texas Holdem and the drawing of video poker. Also note  that the objective of the triple draw poker is to make a low hand, not a “high hand” as in video poker. The best hand in 2-7 triple draw is 2, 3, 4, 5, 7 in any order.



\end{enumerate}

As we can see, these three games have very different rules and objectives with 2-7 triple draw poker being the most complex. In the following section, we describe a unified representation for these games.


\subsection{A Unified Representation for Poker Games}

A key step in our Poker CNN model is to encode the game information into a form which can be processed by the convolution network.  

There are 13 ranks of poker cards and four suits\footnote{c = club, d = diamond, h = heart, s = spade}, so we use a $4\times 13$ sparse binary matrix to represent a card. Only one element in the binary matrix is non-zero. 
In practice, we follow the work in \cite{clark2014teaching} by 
zero-padding each  $4\times 13$ matrix to $17\times 17$ size. Zero padding does not add more information, but it helps the computation with convolutions and max pooling. For a five card hand, we represent it as a $5\times 17\times 17$ tensor.  We also add a full hand layer ($17\times 17$) that is the sum of 5 layers to capture the whole-hand information.

There are a number of advantages to such an encoding strategy: First, a large input creates a good capacity for building convolution layers. Second, the full hand representation makes it easy to model not only the individual cards, but also common poker patterns, such as a pair (two cards of the same rank, which are in the same column) or a flush (five cards of the same suit, which are in the same row) without game-specific card sorting or suit isomorphisms (e.g. AsKd is essentially the same as KhAc). Finally, as we show next, we are able to extend the poker tensor to encode game-state context information that is not measured in cards. 

For multiple round games, 
we would like to keep track of context information such as the number of draw rounds left and the number of chips in the pot. We do so by adding layers to the $6\times 17\times 17$ tensor. To encode whether a draw round is still available, we use a $4\times 13$ matrix with identical elements, e.g. all ``1'' means this draw round is still available. We pad this matrix to $17\times 17$. For 2-7 Triple Draw, there are three draw rounds, therefore we add three $17\times 17$ matrices to encode how many rounds are left. To encode the number of chips in the pot, we add another layer using numerical coding. For example, if the number of chips is 50, we encode it as the 2c card, the smallest card in the deck. If 200, we use 2c\&2d\&2h\&2s to encode it. A pot of size 2800 or more will set all 52 cards' entries to be 1.

We also need to encode the information from the bets made so far. Unlike for a game like chess, the order of the actions (in this case bets) used to get to the current pot size is important. First we use a matrix of either all ``0'' or all ``1'' to encode whether the system is first to bet. Then we use 5 matrices to encode all the bets made in this round, in the correct order. Each matrix is either all ``0'' or all ``1'', corresponding to 5 bits to model a full bet round. We similarly encode previous betting rounds in sets of 5 matrices. 

The final representation for triple draw is a $31\times 17\times 17$ tensor whose entries are either 0 or 1. Table~\ref{table-triple-draw-features} explains encoding matrices for triple draw. To represent Texas Hold'em, we just need to adjust the number private cards to encode the observed public cards, remove draw rounds layer since there is not card drawing by the player. 

To the best of our knowledge, this is the first work to use a 3D tensor based representation for poker game states. 




\begin{table}[b!]
	\caption{Features used as inputs to triple draw CNN }
	\begin{center}
		\begin{tabular}{lrl }
        	Feature & Num. of matrices & Description \\
            \hline 
           	xCards & 5 & Individual private cards \\
            	  & 1 & All cards together \\
            xRound & 3 & Number of draw rounds left \\
            xPotSize & 1 & Number of chips in the pot \\
            xPosition & 1 & Is the player first to bet? \\
            xContext & 5 & Bets made this round \\
            		 & 5 & Bets made previous round \\
                     & 5 & \# Cards kept by player \\
                     & 5 & \# Cards kept by opponent \\
            \hline 	
		\end{tabular}
	\end{center}
	\label{table-triple-draw-features}
\end{table}

\section{Learning}

A poker agent should take two types of actions in the game: drawing cards and placing bets. In the following sections, we describe how to learn these two types of actions.

\subsection{Learning to draw}

We consider the task of learning to make draws as the following: given a hand of poker cards, estimate the return of every possible draw. In a poker game where the player can have five cards, there are $2^5=32$ possible choices per hand, so the machine needs to estimate the gain/loss for each of the possible 32 choices and select the choice with biggest gain. Since the 2-7 Triple Draw game involves betting as well as draws, and Texas Hold'em involves only betting, we use video poker as an example to illustrate our approach to learning to make optimal draws in a poker game.

In many cases, it is easy to heuristically judge whether a hand include a good draw or not. For example, a hand with five cards of the same suit in a row is a straight flush, and will be awarded \$50 in video poker. If a player sees four cards or the same suit in a row and an uncorrelated card, he will give up the last card and hope for a straight flush. To estimate the possibility for their action, we employ Monte Carlo simulation, and average the return. If with 1\% probability he wins \$50 and the other 99\% wins nothing, the estimate return for this choice is $\$50\times 0.01=\$0.5$. 


To train a player for video poker, we  employ Monte Carlo simulation  to generate 250,000 video poker hands. From simulation, we obtain  the  expected gain/loss of each possible draws, for each of these 250,000 hands.

\begin{figure}[t!]	
	\begin{center}		
		\includegraphics[width=8cm]{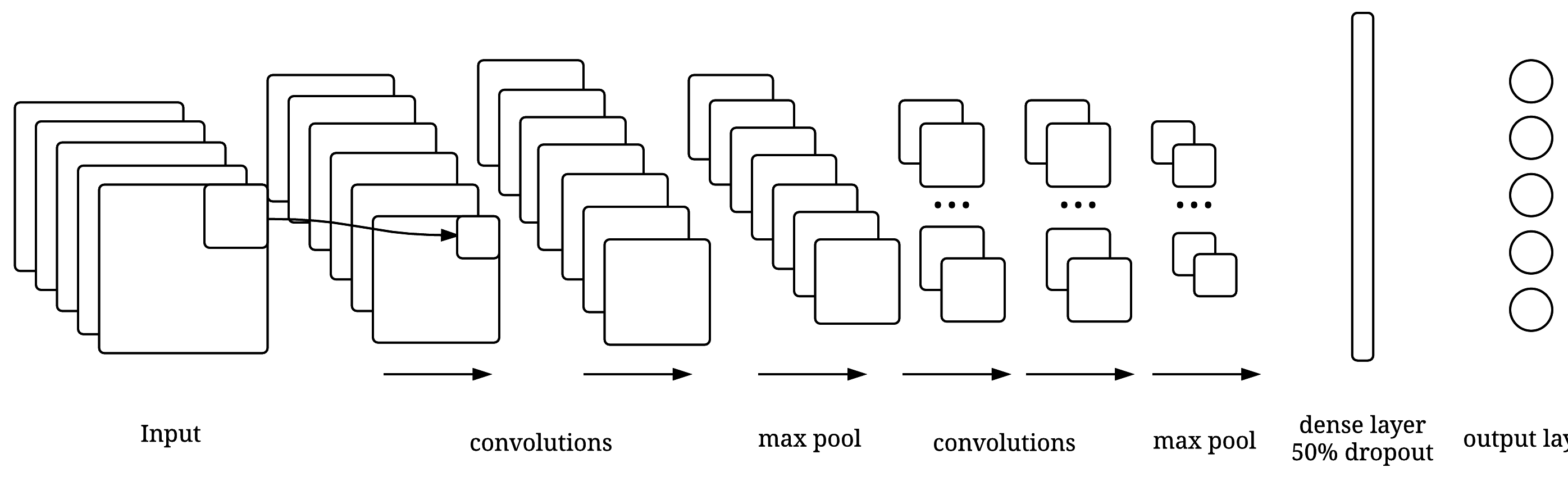}
	\end{center}
	\caption{The structure of our poker-CNN. }
    \label{fig-poker-CNN-details}
\end{figure}

Based on the representation introduced in the previous section (without the betting information), we consider several models for predicting the player action with highest estimated value:

\begin{description}
\item [Fully connected neural network]. Fully connected neural networks, or multi-layer perceptrons, are efficient nonlinear models which can be scaled to a huge amount of training examples. We use a model with two hidden layers with 1032 hidden units in each layer, and the output of last layer is 32 units. We also add a dropout layer before the output to reduce overfitting. 

\item  [Convolutional  network with big filters]. We believe it is useful to learn 2D patterns (colors and ranks) to represent poker. The successful approaches in image  recognition \cite{LeCun-PIEEE-1998}
suggest using convolutional filters to recognize objects in 2D images. Motivated by their work, we introduce a CNN model, named Poker-CNN, which consists of four convolution layers, one maxpool layer, one dense layer, and a dropout layer. The filter size of this model is $5\times 5$. The structure of our network is shown in Figure~\ref{fig-poker-CNN-details}.

\item [Convolutional network with small filters]. Some recent work \cite{Simonyan14c} shows that it can be beneficial to use a deep structure with small filters. Motivated by this work, we use $3\times 3$ filters in the convolutional layers. 
\end{description}
All of our models were implemented using Lasagne \cite{sander_dieleman_2015_27878}.  We did not use a nonlinear  SVM because it would be prohibitively slow with a large number of support vectors. All these models use mean squared error as the loss function.
Nesterov momentum \cite{sutskever2013importance}
is used to optimize the network with
a learning rate of 0.1 and momentum of 0.90.

We can use the same strategy to learn how to make draws in other games.
For 2-7 Triple Draw, simulating the value of 32 possible choices is a bit more complicated, since there are multiple draws in the game. We use depth-first search to simulate three rounds of draws, then average the  return over all of the simulations. Rather than the average results against a payout table, we train the Triple Draw model to predict the odds of winning against a distribution of random starting hands, which also have the ability to make three draws. Thus the ground truth for triple draw simulation is an allin value of 0.0 - 1.0, rather than a dollar value.

The CNN models for 2-7 Triple Draw are similar to those for video poker, except that the input may contain more matrices to encode additional game information, and there are more filters in the convolutional layer. 
For example, the number of filters for video poker is 16  in the first two convolutional layers and 32 in the last two layers. In contrast, the number of filters  for triple draw is 24 and 48, respectively. 

For complex poker games like Texas Hold'em and 2-7 Triple Draw, we also need a model that can learn how to make bets.
Next, we consider the task of choosing when to check,  bet, raise, call or fold a hand. The task of making bets is much harder than making draws, since it is not so simple to produce ground truth through Monte Carlo simulation. In the following sections we will discuss how our model learns to make bets, and how it reacts to an opponent's strategy.

\subsection{Learning to bet}

The fun part in most modern poker games is making bets. In each round of betting, players act in turn, until all players have had a chance to bet, raise, or call a bet. The player has at most five choices: ``bet'' (start to put in chips), ``raise'' (increase the bet), ``call''(match the previous bet made), ``check'' (decline to make the first bet), 
and ``fold'' (give up the hand). There are two things worth pointing out. First, a player need not be honest. He may bet or raise with a bad hand in the hope of confusing his opponents. Second, no player is perfect. Even world-class players make imperfect choices in betting, nor will a player always make the same choice in a given situation. 


Ideally, we would like to use the same network as in Figure~\ref{fig-poker-CNN-details} to make bets. 
Differently from making draws,  we predict the expected values of five output states, corresponding to five betting actions. However, the challenge  is that it is no longer easy to use Monto Carlo simulation to collect the average return of every betting choice. 


To generate training examples for making bets in various game states, we propose to track the results in terms of chips won or lost by this specific betting action. If a player folds, he can not win or lose more chips, so his chip value is \$0. Otherwise, we count how many chips the player won or lost with this action, as compared to folding. This includes chips won or lost by future actions taken during the simulated hand.

We use a full hand simulation to generate training examples for the bets model. At the beginning, we employ two randomized heuristic players to play against each other. A simple but effective heuristic player is to randomly choose an action based on a fix prior probability. 
Our heuristic player chooses a simple strategy of a 45\% bet/raise, 45\% check/call, 10\% fold baseline. Depending on whether it is facing a bet, two or three of these actions are not allowed, and will not be considered. During play, the probability of choosing a betting action is 
adjusted to the current hand's heuristic value\footnote{As estimated by the allin value earlier.}. This adjustment makes sure that the simulated agent very rarely folds a very good hand, while folding a bad hand more often than it would an average hand. Given a game state, we aim to predict the gain/loss of each possible betting action.  To train such a model, we use the  gain/loss of full hands simulation, which is back-propagated after the hand finishes.

We generated 500,000  hands for training. In practice, we first learn the network for making draws, and use the parameter values as initialization for the network of making bets.
As we did for learning to make draws, we use
Nesterov momentum \cite{sutskever2013importance}
for learning to make bets, except that we use a smaller 
learning rate of 0.02 for optimization.
Note that in every training example, the player makes only  one decision, so we only update parameters corresponding to the given action. Since our heuristic player is designed to make balanced decisions, our network will learn to predict the values of each of the five betting actions.

\subsection{Learning from inferior players}

One concern about our learning model is that the training examples are generated based on simple heuristic players, hence these examples are not exhaustive, prone to systemic biases, and not likely to be representative of all game situations. 

We hypothesize that our model will outperform the simple heuristic players from which it learns for two reasons. First, our model is not learning to imitate the actions taken in the heuristic-based matches, but rather it is using these matches to learn the value of the actions taken. In other words, it learns to imitate the actions that lead to positive results. 

Second, the heuristic-based player's stochastic decision-making includes cases where a player bets aggressively with a weak hand (e.g. bluffing). Therefore, the resulting model is able to learn the cases where bluffing is profitable, as well as when to bet with a strong hand.

To further improve the performance of our models, we vary our training examples by letting the new model not only play against itself, but also against the past two iterations of trained models. We find this strategy is very effective. For our final model, we repeated the self-play refining process 8 times, each time with self-play against the current best model, and also previous models. That way, the next version of the model has to play well against all previous versions, not just to exploit the weaknesses of the specific latest model. We will show the resulting improvement of iterative refining in our experiments.

\section{Experiments on Video Poker}


%
%
%
Video poker is a simple game which only requires one to make draws. We will discuss how our model plays on video poker in this section so that it will be easier to understand  more complicated game in the next section.
Table~\ref{video-poker-results} compares the performance of different models for video poker. Note that if a player plays each hand perfectly, the average payout is \$1.000, but few human players actually achieves this result\footnote{That is why the casino always wins}. Top human performance is usually 1\%-2\% worse than perfect \footnote{http://www.videopokerballer.com/articles/10-video-poker-mistakes/}. However since video poker is a simple game, it is possible to develop a heuristic (rule-based) player that returns \$0.90+ on the dollar\footnote{Our heuristic uses four rules. A 25-rule player can return \$0.995 on the dollar. http://wizardofodds.com/games/video-poker/strategy/jacks-or-better/9-6/intermediate/}. 

We compare our models with a perfect player and a heuristic player. It is easy to see that all of our learning models outperform the random action player and the heuristic player. The performance of our best model (\$0.977) is comparable with the strong human player (\$0.98-\$0.99). 

\begin{table}[h!]
	\caption{Poker-CNN's average return in the video poker.}
	\label{video-poker-results}
	\begin{center}
		\begin{tabular}{lcc }
        	model & average return \\
			\hline
            perfect player & \$1.000 \\
            professional player & \$0.98 to \$0.99 \\
            heuristic player &  \$0.916 $\pm$ 0.002 \\
            random action player & \$0.339 $\pm$  0.003 \\
            \hline
            Fully connected network & \$0.960  $\pm$ 0.002 \\
            CNN with $5 \times 5$ filters & \$0.967  $\pm$ 0.002 \\
			CNN with $3\times 3$ filters & \$0.977 $\pm$ 0.002 \\
            \hline
		\end{tabular}
	\end{center}
\end{table}

Table~\ref{video-poker-errors} takes a closer look at the differences in learning models by
comparing the errors in each category. It is interesting to see that CNNs make fewer ``big'' mistakes than DNNs. We believe this is because CNNs view the game knowledge as patterns, i.e., combinations of suits and ranks. Since the search space is very large, the CNN model has a better chance to 
learn the game knowledge by capturing the patterns in 2D space than a DNN might in 1D space, given a fixed number of examples.

\begin{figure}[t!]
	\begin{center}
\includegraphics[width=6cm]{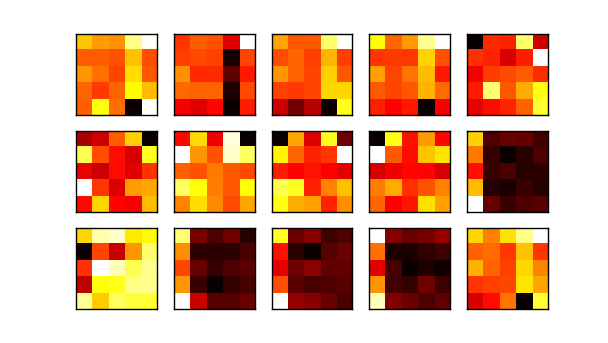}
	\end{center}
	\caption{Filters learned from video poker.}
    \label{fig-video-poker-CNN-filters}
\end{figure}

To demonstrate this, Figure~\ref{fig-video-poker-CNN-filters} shows a sample of the filters learned in the first layer of CNN network. These filters are applied to a $5\times 17\times 17$ tensor, representing the five cards in 2D space. Vertical bars in the image are suggestive of a filter looking for a pattern of cards of the same rank (pairs), while horizontal bars in the image are suggestive of a filter looking for cards of the same suit (flushes).

\begin{table}[t!]
	\caption{Comparing errors for video poker models}
	\label{video-poker-errors}
    \small{
	\begin{center}
		\begin{tabular}{lccccc }
        	model & negligentable & tiny & small  & big  & huge  \\
            	  & $<$\$0.005 & $<$\$0.08 & $<$\$0.25 & $<$\$1.0 & \$1.0+ \\
			\hline
            heuristic & 1610 & 53 & 1081 & 256 & 0 \\
            \hline
            DNN & 2327 & 348 & 242 & 83 & 23  \\
            CNN $5 \times 5$  & 2322 & 413  & 151  & 114 & 8 \\
			CNN $3\times 3$  & 2467  & 353  & 131  & 49  & 8 \\
            \hline
		\end{tabular}
	\end{center}
    }
\end{table}



The experiments on video poker suggest that the CNN is able to learn the patterns of what constitutes a good poker hand, and estimate the value of each possible move, thus deducing the move with the highest expected payout. 

\section{Experiments on Texas Hold'em \& Triple Draw}

We have discussed the simple video poker game in previous Sections. Now we will discuss more complicated games: heads-up limit Texas Hold'em and heads-up limit 2-7 Triple Draw.

We first generate 500,000 samples from which our model learns how to make draws (no draws for Texas Hold'em). Then we let two heuristic players\footnote{Our heuristic player is based on each hand's allin value against a random opponent.} play against each other for 100,000 hands and learn how to make bets based on those hand results.

Since making bets is difficult, we let the model play against itself, train on those results, and repeat this self-play cycle a number of times. As mentioned before, each round of self-play consists of the latest CNN model playing against the previous three models 
to avoid over-optimizing for the current best model's weaknesses. 

For each training iteration, we generate 100,000 poker hands, and from those, train on 700,000 betting events. Our final Texas Hold'em model was trained for 8 self-play epochs, while the Triple Draw model was trained for 20 self-play epochs.

\subsection{Playing Texas hold'em}

To evaluate the performance of poker player, we compare our best poker-CNN with the following models: 
\begin{itemize}
	\item ``Random player'' from ACPC. \footnote{http://www.computerpokercompetition.org/}
    \item The heuristic player, from which our model is trained.
	\item An open source implementation of a CFR\footnote{https://github.com/rggibson/open-pure-cfr} player. 
\end{itemize}

Table~\ref{table-holdem-cnn-vs-others} summarizes the results. 
The first notable result is that our  model outperforms the heuristic player from which it learns. This supports our hypothesis that the poker playing model can outperform its training data. 

Our model also outperforms the random player and the open source CFR player. The open source CFR~\cite{richard-gibson-thesis-phd14} computes a Nash equilibrium solution using the ``Pure CFR" algorithm. That is the only public implementation we can find for a version of the CFR algorithm used by the ACPC Limit Holdem champions. However, the limitation of this implementation is that it  abstracts all poker hands as one bucket. Although it clearly illustrates the idea of CFR, we recognize that its performance is not close to the best CFR implementation to date. To remedy this, we asked a former professional poker player to compete against our model. He won  $+21.1 \pm 30.5$ over about 500 hands. 
From this result, we can see that our model is competitive against a human expert, despite being trained without any Texas Hold'em specific domain knowledge. 


\begin{table}[t!]
	\caption{Players' earnings when playing against Poker-CNN in heads up limit Texas hold'em, with \$50-\$100 blinds. The $\pm$ amount indicates error bars for statistical significance.}
	\label{table-holdem-cnn-vs-others}
	\begin{center}
		\begin{tabular}{lccl }
        	 Player & Player earnings & \# hands   \\
            \hline
            ACPC sample player & -\$90.9 $\pm$7.0 & 10000  \\
            Heuristic player & -\$29.3 $\pm$5.6 & 10785  \\
            	\hline
            CFR-1 & -\$93.2 $\pm$7.0 & 10000 \\
            Professional human player & +\$21.1$\pm$30.5 & 527 \\
                \hline
		\end{tabular}
	\end{center}
\end{table}



\subsection{Playing 2-7 triple draw poker}
\label{sec:triple}
Triple draw is more complicated than Texas Hold'em, and there is no existing solution for 2-7 triple draw. 
To validate our performance, we compare the heuristic player, CNN-1 (trained by watching two heuristic players playing), the final CNN player (after 20 iterations of self-play and retraining), and a DNN, trained on the results of the  Poker-CNN model. The results show that
CNN-1 beats the heuristic model by large margin, despite being trained directly on its hands. Iterative training leads to improvement, as 
our model significant outperforms both the heuristic model and CNN-1. Our model outperforms the two-layer DNN  with a gain of $\$66.2$, despite the DNN being trained on the CNN hands, from the latest model.

\begin{table}[t!]
	\caption{Different models play each other in 2-7 Triple Draw. Results over 60,000+ hands, significant within $\pm$\$3.0 per hand.
}
	\label{triple-draw-results-matrix}
	\begin{center}
		\begin{tabular}{lrrrr }
        	model & heuristic & CNN-1 & Poker-CNN & DNN \\
            \hline
			Heuristic & 0 & -\$99.5 & -\$61.7 & -\$29.4  
            \\
            CNN1 & +\$99.5 & 0 & -\$73.3 & -\$54.9
            \\
            Poker-CNN & +\$61.7 & +\$73.3 & 0 & +\$66.2 
            \\
            DNN & +\$29.4 & +\$54.9 & -\$66.2 & 0 \\
            \hline
		\end{tabular}
	\end{center}
\end{table}


Since there is no publicly available third-party 2-7 triple draw AI to compete against, Poker-CNN model was pitted against human players.  We recruited a former professional poker player (named as ``expert") and a  world champion poker player (named as ``champion") to play against our poker-CNN model. 

Evaluating against an expert human player takes considerably longer than comparing two models, so our experts only played 500 hands against the model.  As Table~\ref{table-triple-draw-cnn-vs-expert} shows, our model  was leading the expert after 100 hands, and broke even at the halfway point. The human expert player noted that the CNN made a few regular mistakes and was able to recognize these after a while, thus boosting his win rate. He also said that the machine was getting very lucky hands early. It would be interesting to see if these weaknesses disappear with more training.  By the end of the match, the human was playing significantly better than the model. The play against world champion shows similar patterns, where  the world champion improves his ability to exploit the CNN. The world champion's final performance against the Poker-CNN is $\$97.6\pm30.2$, but praised our system after the game, on being able to play such a complicated game. We think the results  demonstrate that a top human player is still better than our model, but we believe that the machine is catching up quickly. 



\begin{table}[t!]
	\caption{Poker-CNN vs human experts for 2-7 Triple draw.}
	\label{table-triple-draw-cnn-vs-expert}
	\begin{center}
		\begin{tabular}{ccc }
        	 \# hands & expert vs CNN & champion vs CNN \\
            \hline
            100 & -\$14.7 $\pm$58.5 & +\$64.0 $\pm$67.5  \\
			200  & +\$4.5 $\pm$39.8 & +\$76.2 $\pm$47.7 \\
            500 & +\$44.6 $\pm$24.2 & +\$97.6 $\pm$30.2 \\
            	\hline
               
		\end{tabular}
	\end{center}
\end{table}





\section{Discussion}

The overreaching goal of this work is to build a strong poker playing system, which is capable of learning many different poker games, and can train itself to be competitive with professional human players without the benefit of high-quality training data. 

However, we also wanted to show a process that is as straightforward and domain-independent as possible. Hence, we did not apply game-specific transformations to the inputs, such as sorting the cards into canonical forms, or augmenting the training data with examples that are mathematically equivalent within the specific game type\footnote{In Texas Hold'em, there are $52*51=2652$ unsorted preflop private card combinations, but these form just 169 classes of  equivalent preflop hands.}. We should try the former, training an evaluating hands in a canonical representation. It is notable that human experts tend to think of poker decisions in their canonical form, and this is how hands are typically described in strategy literature\footnote{A hand like KsAc is described as "Ace, King off-suit," regardless of order or specific suit.}. \cite{waugh2013fast} and others have developed efficient libraries for mapping Texas Hold'em hands to their canonical form. It should be possible to extend these libraries to other poker variants on a per-game basis. 

Within our existing training process, there are also other optimizations that we should try, including sampling from generated training data, deeper network,  more sophisticated strategies of adaptively adjusting  learning rates, and etc. We feel these may not make a significant difference to the system's performance, but it will be definitely worth exploring.

Inspired by the success of \cite{mnih2015human}, we set out to teach a deep Q-learning network to play poker from batched self-play. However, it turned out that we could create a strong CNN player by training directly on hand results, instead of training on intermediate values. This is possible because the move sequences in poker are short, unlike those for go, backgammon or chess, and that many moves lead immediately to a terminal state. It will be interesting to see how much better the system performs, if asked to predict the value of its next game state, rather than the final result.

Lastly, in building our systems, we've completely avoided searching for the best move in a specific game state, or explicitly solving for a Nash equilibrium game strategy. The upside of our approach is that we generate a game strategy, directly from self-play, without needing to model the game. As a bonus, this means that our method can be used to learn a strategy that exploits a specific player, or a commonly occurring game situation, even if it means getting away from an equilibrium solution. We should explore this further.

We acknowledge that broad-ranging solutions for complicated poker games will likely combine the data driven approach, along with search and expert knowledge modeling. It is our understanding that the No Limit Texas Hold'em system that competed to a statistical draw with the world's best human players \cite{rutkin2015ai}, already uses such a hybrid approach. Recent works \cite{DBLP:conf/icml/HeinrichLS15} have also shown that it is possible to closely approximate a Nash equilibrium for heads-up Limit Texas Hold'em, using a Monte Carlo search tree combined with reinforcement learning.  

It would be interesting to see how these different approaches to the same underlying problem, could be combined to create strong player models for very complicated poker games, such as competing against six strong independent players in ten different poker games\footnote{Known as "mixed" or "rotation" games by professional poker players.}.

At the very least, we would like to fully implement an up to date version of the CFR algorithm for heads up Limit Hold'em, which has has been shown to play within \$6.5 per hand of a perfectly unexploitable player, on our \$50-\$100 blinds scale, even when limited to 10,000 abstraction buckets, which a single machine can reasonably handle \cite{johanson2013evaluating}. It would be interesting to see where Poker-CNN places on the CFR exploitability spectrum, as well as how much Poker-CNN could improve by training on hands against a strong CFR opponent.

\section{Conclusion and future work}

We have presented Poker-CNN, a deep convolutional neural network based poker playing system. It learns from  a general input representation for poker games, and it produces competitive drawing and betting models for three different poker variants. We believe that it is possible to extend our approach to learn more poker games, as well as to improve the ability of this model to compete on an even footing with the best available equilibrium-learning models, and against professional poker players. 

Perhaps not far from today, it will be possible to train a poker model that can play any poker game spread at the casino, at a level competitive with the players who make a living playing these games. We would like to see Poker-CNN make a contribution toward this future.

\section{Acknowledgments}

We would like to thank professional poker players Ralph "Rep" Porter, Randy Ohel and Neehar Banerji for playing against Poker-CNN, as well as for invaluable insights into the system's performance. Thanks also to Eric Jackson and Richard Gibson, for sharing their experience with CFR methods for Limit and No Limit Texas Hold'em, as well as for sharing some of their code with us. 

\bibliographystyle{aaai}
\bibliography{cnnBib,pokerBib}

\end{document}